\documentclass[runningheads]{llncs}
\usepackage{graphicx}
\usepackage{booktabs}       
\usepackage{amsfonts,amsmath,amssymb}       
\usepackage{nicefrac}       
\usepackage{microtype}      
\usepackage{lipsum}
\usepackage{epsfig}

\newcommand{\argmin}{\operatorname*{argmin}}
\newcommand{\argmax}{\operatorname*{argmax}}

\newcommand{\given}{\, | \,}

	\newcommand*{\defeq}{=}

\newcommand{\Prob}{P}

\renewcommand{\vec}[1]{\boldsymbol{#1}}
\newcommand{\cX}{\mathcal{X}}
\newcommand{\cY}{\mathcal{Y}}
\newcommand{\cH}{\mathcal{H}}
\newcommand{\cD}{\mathcal{D}}

\newcommand{\fromto}{\longrightarrow}

\usepackage{ifthen}
\newboolean{commentsoff}
\setboolean{commentsoff}{true}
\usepackage{todonotes}

\newcommand{\todofm}[1]{\ifthenelse{\boolean{commentsoff}}{}{\todo[inline,color=cyan]{FM: #1}}}
\newcommand{\todoat}[1]{\ifthenelse{\boolean{commentsoff}}{}{\todo[inline,color=yellow]{AT: #1}}}
\newcommand{\todoeh}[1]{\ifthenelse{\boolean{commentsoff}}{}{\todo[inline,color=lime]{EH: #1}}}

\begin{document}
\title{Automated Machine Learning, Bounded Rationality, and Rational Metareasoning}
\titlerunning{AutoML, Bounded Rationality, and Rational Metareasoning}
%
\author{Eyke H{\"u}llermeier\inst{1}\orcidID{0000-0002-9944-4108} \and
Felix Mohr\inst{2}\orcidID{0000-0002-9293-2424} \and
Alexander Tornede\inst{3}\orcidID{0000-0002-2415-2186}\and \\
Marcel Wever\inst{3}\orcidID{0000-0001-9782-6818}}
\authorrunning{E.\ H{\"u}llermeier, F.\ Mohr, A.\ Tornede, M.\ Wever}
%
\institute{LMU Munich, Munich, Germany \\
\email{eyke@lmu.de} \and
Universidad de La Sabana, Ch\'{i}a, Cundinamarca, Colombia
\email{felix.mohr@unisabana.edu.co} \and
Paderborn University, Paderborn, Germany\\
\email{\{alexander.tornede, marcel.wever\}@upb.de}}
\maketitle              
\begin{abstract}
The notion of bounded rationality originated from the insight that perfectly rational behavior cannot be realized by agents with limited cognitive or computational resources. Research on bounded rationality, mainly initiated by Herbert Simon,  has a longstanding tradition in economics and the social sciences, but also plays a major role in modern AI and intelligent agent design. Taking actions under bounded resources requires an agent to reflect on how to use these resources in an optimal way\,---\,hence, to reason and make decisions on a meta-level. In this paper, we will look at automated machine learning (AutoML) and related problems from the perspective of bounded rationality, essentially viewing an AutoML tool as an agent that has to train a model on a given set of data, and the search for a good way of doing so (a suitable ``ML pipeline'') as deliberation on a meta-level. 

\end{abstract}

\section{Introduction}

The notion of \emph{rationality} and rational decision making is at the core of economics and the behavioral sciences, and has more recently become a guiding principle of modern artificial intelligence \cite{russ_ai}. Various attempts at formalizing perfect rationality have been made in the realm of mathematical decision theory, with Savage's subjective expected utility theory \cite{sava_tf} as the arguably most prominent representative.
Yet, as noted by Herbert Simon, perfect rationality is difficult to achieve in practice, due to cognitive and computational limitations \cite{simo_ab}: Even if a rational decision or action does exist, limited resources in terms of time and computational power may prevent the agent from finding it. As an alternative to perfect rationality, Simon proposed the concept of  \emph{bounded rationality} and, in this regard, introduced the notion of ``satisficing'' \cite{simo_ab55}. 

\todofm{Gibt's diese Referenzenvon Savage und Simon irgendwo? Das sind ja zwei Bücher, die ich nicht so ohne weiteres bekommen konnte. Auch die nachfolgende Referenz auf Good et al. konnte ich nicht auflösen.}

An agent that must act and make decisions under bounded resources also needs to ``think'' about how to use these resources in an optimal way. This gives rise to another decision problem, namely a decision problem on a \emph{meta-level}, which again causes costs and must be solved under bounded resources.  The point that the cost of decision making must also be taken into account was made by several authors. For example, as an early proponent of this idea, Irving Good coined the term ``type II rationality'' \cite{good_ts71}. What a bounded rational agent should naturally aim for is an optimal (resource-bounded) solution to an \emph{augmented} problem, consisting of the original problem (on the \emph{object-level}) and the ``deliberation'' on the meta-level, i.e., a sequence of computations plus an action in the end. 

The goal of this paper is to establish a connection between two branches of AI, namely previous research on bounded rationality and more recent research on automated machine learning (AutoML). More specifically, we will look at automated machine learning and related problems from the perspective of bounded rationality, essentially viewing an AutoML tool as an agent that has to train a model on a given set of data, and the search for a good way of doing so (a suitable ``ML pipeline'') as deliberation on a meta-level. We believe that this perspective sheds new light on existing AutoML methods and may serve as a guide for future research in this active field of research. 


\section{Types of Rationality}

Going beyond rationality and bounded rationality as mere conceptions and formal ideas, various computational models have been developed in AI for the purpose of practical problem-solving. 
In this regard, Russell \cite{russ_ra16} makes a distinction between four types of rationality.


\paragraph{Perfect Rationality.}
A perfectly rational agent takes the optimal action in every state of the environment $E$ it is acting in, i.e., it acts according to the optimal agent function
$$
f_{opt} = \argmax_{f} V(f, E , U) \, .
$$
Here, optimality refers to the expected value $V$ (assuming a probability distribution on $E$) of a performance measure $U$ on sequences of environment states when the agent takes actions as prescribed by the function $f$.

\paragraph{Calculative Rationality.}

While perfect rationality ``merely'' requires the existence of an optimal agent function $f_{opt}$, calculative rationality additionally takes computational aspects into account. An agent that exhibits calculative rationality is provably able to compute\,---\,through a suitable program executed on a certain machine\,---\,the optimal actions in every state. Such computations must be possible at least in principle, even if the costs might be arbitrarily high. In other words, calculative rationality stipulates requirements for computability but not for complexity.



\paragraph{Meta-Level Rationality.}

The notion of metareasoning is based on the aforementioned distinction between an \emph{object-level}, on which the actual problem is solved, and a \emph{meta-level} of the problem-solving process (and additionally a \emph{ground-level} on which the actions are eventually executed). It refers to the idea of monitoring and controlling the agent's object-level deliberation, i.e., the way in which it solves the actual (decision) problem. Thus, metareasoning stipulates a specific architecture (design philosophy) of an agent program, dividing it into the object-level and the meta-level.

Optimal metareasoning, also known as rational metareasoning \cite{horv_rm89} or metalevel rationality \cite{russ_ra95}, seeks to maximize the overall performance given the agent's object-level deliberation capabilities. At the core of optimal metareasoning is the notion of \emph{value of computation}, which is closely related to the theory of information value also known from Bayesian decision theory \cite{howa_iv66}: Computations on the meta-level are perceived as actions that may change the agent's informational state and hence impact and possibly improve its decisions on the object-level. The problem of metareasoning can therefore be tackled by decision theory, meaning that computations should be selected based on their expected utility \cite{russ_ra97}.  


Zilberstein \cite{zilb_ma08} considers an agent to be bounded rational when its metareasoning component is optimal, which (informally) means that ``given a particular object-level deliberation model, we look for the best possible way to control it so as to optimize the actual ground-level performance of the agent.'' In the literature, several frameworks in line with this notion of bounded rationality have been proposed, including anytime algorithms \cite{dean_aa88}, algorithm portfolios \cite{petr_lp06}, and contract algorithms \cite{zilb_os03}. 


\todofm{Bin hier nicht ganz sicher, ob ich das so teile. Der Satz ``Because meta-level reasoning comes with computational costs, it cannot be solved optimally'' ist so glaube ich nicht richtig. Die Existenz von Rechenkosten alleine heißt ja noch nicht gleich, dass ein Problem nicht effizient gelöst werden kann. Z.B. brauchst du n log n um eine Liste zu sortieren. Das heißt es gibt Kosten, aber wir sind ja noch in P.
Es ist für mich nicht unmittelbar selbstverständlich, dass das meta-Problem auch NP-schwer sein muss, nur weil das object-level problem NP-schwer ist.
Ich glaube es ist recht klar, dass nur weil ich schnell und effizient sagen kann, wie ich meine Ressourcen optimal verteilen will, dass ich dadurch noch nicht das Problem selbst gelöst habe; wäre jedenfalls meine Vermutung.
Das ist wichtig, denn wenn wir das meta-Problem effizient lösen können, dann gibt es auch keinen infiniten Regress.}

\paragraph{Bounded Optimality.}

While the previous notions of rationality refer to the optimality of individual actions or computations, the notion of bounded optimality shifts the attention to entire programs implementing the behavior of an agent. Thus, bounded optimality takes into account that, given a computational device (machine) $M$, not all agent programs are actually feasible. Instead, only a subset of agent functions can be produced by running a program $P$ on $M$. 

According to \cite{russ_dt}, an agent exhibits bounded optimality ``if its program is a solution to the constraint optimization problem presented by its architecture.'' Naturally, then, the task is to optimize over programs rather than actions: Given a machine $M$ and an environment $E$, a program $P_{opt}$ is bounded optimal if running that program yields an expected utility that is higher than for any other program (in a given program class):
$$
P_{opt} = \argmax_{P \in \mathcal{P}_M} V \big( f(P,M) , E , U \big) \, ,
$$
where $\mathcal{P}_M$ is the set of programs that can be run on machine $M$, and $f(P,M)$ the agent function produced by running program $P$ on machine $M$. 


\section{Machine Learning}

In the basic setting of supervised learning, a learner $A$ is given access to a set of training data $ \mathcal{D} = \big\{ (\vec{x}_i , y_i ) \}_{i=1}^N$,
where $\mathcal{X}$ is an instance space and $\mathcal{Y}$ the set of outcomes that can be associated with an instance. Typically, the training examples $(\vec{x}_i , y_i)$ are assumed to be independent and identically distributed (i.i.d.) according to some unknown probability measure $\Prob$ on $\mathcal{X} \times \mathcal{Y}$. Given a \emph{hypothesis space} $\mathcal{H}$ (consisting of hypotheses $h:\, \cX \fromto \cY$ mapping instances $\vec{x}$ to outcomes $y$) and a loss function $\ell: \, \mathcal{Y} \times \mathcal{Y} \longrightarrow \mathbb{R}$, the goal of the learner is to induce a hypothesis $h = A(\cD) \in \mathcal{H}$ with low risk (expected loss)
\begin{equation}\label{eq:risk}
R(h) \defeq \int_{\cX \times \cY} \ell( h(\vec{x}) , y) \, d \, \Prob(\vec{x} , y) \enspace .
\end{equation}
Looking at this setting from the perspective of decision making, the learner is now playing the role of the agent (decision maker) on the object-level. The task of the agent is to select a hypothesis $h$, which corresponds to an action. Thus, the action space of the agent is given by the hypothesis space $\mathcal{H}$, and the agent function maps data sets $\mathcal{D}$ to hypotheses $h$. 
 Once a hypothesis $h$ has been chosen, it can be used to make predictions $h(\vec{x})$ for query instances $\vec{x}$. Such predictions can be seen as actions on the ground-level, and the \emph{predictor} as another agent with action space $\mathcal{Y}$ in the environment $\mathcal{X}$. The function of this agent maps instances to predictions.



The perfectly rational predictor is the pointwise Bayes predictor that chooses
\begin{equation}\label{eq:pbayespred}
f^*(\vec{x}) \defeq \argmin_{\hat{y} \in \cY} \mathbb{E}\big( \ell( y, \hat{y}) \given \vec{x} \big) \, ,
\end{equation}
for every $\vec{x} \in \cX$, where $\mathbb{E}$ denotes the expected value operator. By selecting the prediction that leads to the minimum expected loss given $\vec{x}$, separately for every $\vec{x}$, this agent optimizes individual decisions. 
However, the pointwise Bayes predictor (\ref{eq:pbayespred}) might not be contained in the hypothesis space $\mathcal{H}$, which means that it cannot be represented by the agent. The Bayes predictor is given by  
\begin{equation}\label{eq:bayespred}
h^* \defeq \argmin_{h \in \cH} R(h) \, ,
\end{equation} 
i.e., by the hypothesis with minimal risk within the hypothesis space $\mathcal{H}$. Adopting this hypothesis somehow fits with the idea of bounded optimality, although here the boundedness refers to limited expressive power rather than limited computational resources: Although $h^*$ may not produce the best decision $h^*(\vec{x})$ in every state $\vec{x}$, it performs best on average.

What can be said about the rationality of the learner on the object-level? Recall that the task of the learner is model induction: It takes a data set $\mathcal{D}$ as input and produces a hypothesis $h \in \mathcal{H}$ as output. Obviously, the question of rationality is difficult to answer for this agent, as it comes down to asking for ``rational induction'', i.e., for the best way of solving the problem of inductive inference. Clearly, there is no consensus in the literature, and different answers will be given by different statistical schools. 
Bayesian inference (learning), for example, could be viewed as ``rational'' in so far as it can be justified axiomatically. 


The idea of boundedness and bounded rationality does not seem to play a significant role in standard machine learning.
On the level of the \emph{problem definition}, i.e., the specification of the learning problem, computational or other resource constraints (e.g., time limits) are neither very strict nor made very explicit in common ML settings. 
On the \emph{algorithmic} level, concrete agent programs (learning algorithms) for supervised machine learning typically follow a fixed course of actions with very few choice points at runtime, leaving little room for ``deliberation'' in the sense of monitoring and control on a meta-level.

\section{Automated Machine Learning}


In AutoML \cite{hutter2019automated,he2021automl,wever2021automl,zoller2021benchmark}, the learner is no longer restricted to a pre-defined learning model $(\mathcal{H}, A)$, consisting of a hypothesis space and a learning algorithm. Instead, it can choose among a broad spectrum of learning algorithms $A \in \mathcal{A}$, each associated with an underlying hypothesis space $\mathcal{H}_A$. Here, an algorithm $A$ may also be an ML-pipeline with a complex structure, being composed of several simpler algorithms. Moreover, every algorithm normally has certain hyperparameters $\omega$ that need to be set. A learning model is now a triplet $M = (\mathcal{H}_A , A, \omega)$.

The goal of the learner is to find 
\begin{equation}\label{eq:automl_problem}
   A^* \in  \arg\min_{A \in \mathcal{A}, \omega \in \Omega(A)} R(A_\omega(\mathcal{D}))  \, ,
\end{equation}
where $\Omega(A)$ is the space of hyperparameter configurations for algorithm $A$. The optimal algorithm $A^*$ can then be used to induce $h^* = A^*(\mathcal{D}) \in \mathcal{H} = \bigcup_{A\in\mathcal{A}} \mathcal{H}_A$. To distinguish the AutoML learner from an ``object-level'' learning algorithm $A_\omega$, we also refer to the latter as an \emph{inducer}. 

Instead of applying an inducer $A$ in a pre-defined manner, solving (\ref{eq:automl_problem}) now requires ``deliberation'' of the learner, namely reflection about which inducer might be best for the data $\mathcal{D}$. We can now distinguish between computations on the object-level and computations on a meta-level: While the former are related to fitting a model to the data $\mathcal{D}$ by running a concrete inducer $A_\omega$, the latter are related to actually finding the inducer by selecting $A \in \mathcal{A}$ and $\omega \in \Omega(A)$. 

Resource constraints (typically specified in terms of an upper time limit) and metareasoning now become essential, simply because of the size and complexity of decisions (the different learning models $M$) available to the learner on the meta-level. The majority of AutoML tools \cite{thornton2013auto,olsonM16,feurer2019auto,mohrWH18mlplan,feurer2018practical,wever2019automating,autoband,tornede2020automl,tornedeTWH21} 
search this space by alternating between two steps: they \emph{generate} a candidate inducer $A_\omega$ and then \emph{evaluate} its quality. To this end, $A_\omega$ is given access to the training data $\mathcal{D}$, and the quality of a hypothesis produced by $A_\omega$ is approximated by the empirical error of $h = A_\omega(\mathcal{D} \setminus \mathcal{D}_{val})$ on the validation part of the data, $\mathcal{D}_{val}$ (split from $\mathcal{D}$ in order to obtain an unbiased estimate of the test error).

As can be seen, deliberation of the learner essentially comes down to anticipating the consequences of a decision $M=(\mathcal{H}_A, A,\omega)$, which in turn is costly as it involves the execution of an inducer with a subsequent empirical evaluation of the hypothesis generated. Time limits are therefore needed and well justified. In addition to time, one may argue that an AutoML learner has to reflect on the optimal use of another limited resource, namely the data. For example, how should the data be split into training and validation data?

Let us briefly mention some examples of meta-level deliberation in the context of AutoML and the related problem of hyperparameter optimization \cite{bergstra2011algorithms,feurer2019hyperparameter}:
The order in which inducers are evaluated plays a major role in returning the best possible inducers in the shortest possible time.
To this end, for example, auto-sklearn \cite{feurer2019auto} employs a warm starting mechanism that first evaluates inducers that have already performed well on previously considered data sets. A more recent version of auto-sklearn also reasons about which inducers are considered at all and the procedure used to evaluate them \cite{feurer2018practical}. Another example of metareasoning can be found in ML-Plan \cite{mohrWH18mlplan}, which divides the overall AutoML process into a search phase for exploring new inducers and a subsequent selection phase, in which a set of the most promising inducers are reexamined to make a final decision. Depending on the given data set and the already explored inducers, ML-Plan reasons about when to end the search phase and proceed with the selection phase. 
In the related problem of hyperparameter optimization, often the data presented to the inducer for evaluation is budgeted to allocate computational resources to more promising candidates \cite{jamiesonT16_sh,liJDRT17hyperband}. Other approaches incorporate the cost for observing improvements into the optimization process itself \cite{snoek2012practical,swersky2013multitaskbo}.

\section{Conclusion}

We (re-)considered automated machine learning (AutoML) from the perspective of bounded rationality, adopting the view of an AutoML tool as an agent that trains a model on a given set of data\,---\,searching for a good way of doing so can then be seen as deliberation on a meta-level. The main motivation for adopting this perspective is twofold: First, to shed light on existing AutoML methods, and second, to inspire new approaches based on established principles of rational metareasoning and bounded optimality.

Obviously, these goals are beyond what could be accomplished in this short paper. As a next step, we plan a closer inspection of concrete AutoML scenarios and existing methods used in these scenarios. This includes
algorithm selection \cite{kotthoff2016algorithm,kerschke2019automated,tornede2020run2survive},
meta algorithm selection \cite{tornede2020towards,tornede2021algorithm},
extreme algorithm selection \cite{tornede2019algorithm,tornede2020extreme},
hyperparameter optimization \cite{bergstra2011algorithms,feurer2019hyperparameter},
combined algorithm selection and hyperparameter optimization  \cite{hutter2019automated,thornton2013auto,feurer2019auto},
algorithm configuration  \cite{hutter2009paramils,ansotegui2009gender,hutter2011sequential},
and dynamic algorithm configuration  \cite{biedenkapp2020dynamic,eimer2021dacbench,speck2021learning,shala2020learning}. 
An interesting question is to what extent existing methods can be seen as bounded rational, how they realize metareasoning, and whether they perhaps even do so in a provably optimal manner. Our impression is that this is not the case, also because the meta-perspective is not explicitly adopted. Therefore, we believe that there is indeed a great potential, not only to better understand existing AutoML tools, but also to improve them through metareasoning techniques and principles of bounded optimality.  

\bibliographystyle{splncs04}
\bibliography{ref-BR.bib}

\end{document}